\documentclass[11pt]{article}

\usepackage[preprint]{acl}

\usepackage{times}
\usepackage{latexsym}

\usepackage{lipsum}
\usepackage{todonotes}
\usepackage{enumitem}
\usepackage{tabularx}
\usepackage{bm,amsfonts,amsmath,dsfont,amssymb}
\usepackage{microtype}
\usepackage{subcaption}
\usepackage{graphicx}
\usepackage{multirow,multicol}
\usepackage{makecell}
\usepackage{booktabs}
\usepackage{dashrule}
\usepackage{threeparttable}
\usepackage[most]{tcolorbox}
\usepackage[table]{xcolor}
\usepackage{colortbl}
\usepackage{xspace}
\usepackage{xfp}
\usepackage[normalem]{ulem}

\definecolor{mygreen}{HTML}{4DAF4A}
\definecolor{myred}{HTML}{E41A1C}

\definecolor{valenceblue}{HTML}{134C87}
\definecolor{valencered}{HTML}{991027}

\newcolumntype{Y}{>{\centering\arraybackslash}X}

\definecolor{1st}{HTML}{EBF5FF}

\newcommand{\first}[1]{%
  \begingroup
  \setlength{\fboxsep}{1pt}
  \colorbox{1st}{%
    \makebox[\dimexpr\linewidth-2\fboxsep-4pt\relax][c]{%
      \rule[-0.3ex]{0pt}{2.0ex}\textbf{#1}%
    }%
  }%
  \endgroup
}

\newcommand{\gain}[2]{%
    {\setlength{\fboxsep}{0.2pt}%
    \scriptsize\colorbox{1st!#1}{\strut #2}}%
}

\newcommand{\loss}[2]{%
    {\setlength{\fboxsep}{0.2pt}%
    \scriptsize\colorbox{myred!#1}{\strut #2}}%
}

\newcommand{\circnum}[1]{
\tikz[baseline=-0.75ex]{
\node[
    circle,
    fill=black,
    text=white,
    font=\bfseries\scriptsize,
    inner sep=0.8pt
] (n) {#1};
}}

\newcommand{\method}{AtmosERC\xspace}

\usepackage[T1]{fontenc}

\usepackage[utf8]{inputenc}

\usepackage{microtype}

\usepackage{inconsolata}

\usepackage{graphicx}

\title{\method: Modeling Dialogue-Level Affective Atmosphere\\for Emotion Recognition in Conversation}

\author{
    Weijie Feng,
    Tongwei Zhang,
    Binbin Liu,
    Zhiyong Cheng\\
    Hefei University of Technology\\
    wjfeng@hfut.edu.cn,
    tongweizhang@mail.hfut.edu.cn,\\
    binbin.liu@hfut.edu.cn,
    jason.zy.cheng@gmail.com
}

\begin{document}
\maketitle

\makeatletter
\def\input@path{{contents/}}
\makeatother
\begin{abstract}

Emotion Recognition in Conversation (ERC) plays a key role in developing emotionally responsive conversational systems, and contextual modeling is central to this task.
Several studies have developed contextual models that capture dialogue history, speaker interactions, and emotion dynamics to support emotion prediction.
However, the specific role of affective atmosphere, a latent, conversation-level emotional tendency shaped by conversational content and speaker interactions, remains insufficiently understood.
This paper investigates what role atmosphere plays in conversational emotion understanding.
We first propose \method, a graph-based atmosphere extractor that utilizes conversational context and speaker interactions to produce a compact representation of affective atmosphere.
We then evaluate two downstream uses of the extracted atmosphere: initializing a lightweight ERC model and providing emotion cues for LLM-based ERC.
Experiments on three widely used ERC benchmarks show that the estimated atmosphere improves emotion prediction across both applications and supports more reliable predictions when local emotions deviate from the overall affective tendency.

\end{abstract}

\section{Introduction}

Emotion Recognition in Conversation (ERC) plays an important role in developing conversational systems that respond appropriately to users' emotional states~\citep{poria2019a,shou2026}.
Extensive research has shown that such emotional understanding depends on interpreting utterances in relation to conversation history and speaker interactions, making contextual modeling central to ERC~\citep{ghosal2019,xue2026}.
However, conversation context combines heterogeneous information, including history events, sociocultural backgrounds, and speaker interactions.
Among these, group-level affective atmosphere that emerges through interpersonal interaction may be particularly relevant to ERC~\citep{barsade2015,jakobsadeh2025}.
While some research has been carried out on global context modeling~\cite{wang2020}, the specific contribution of affective atmosphere to utterance-level emotion recognition warrants systematic investigation.

\begin{figure}[!t]
    \centering
    \includegraphics[width=\columnwidth]{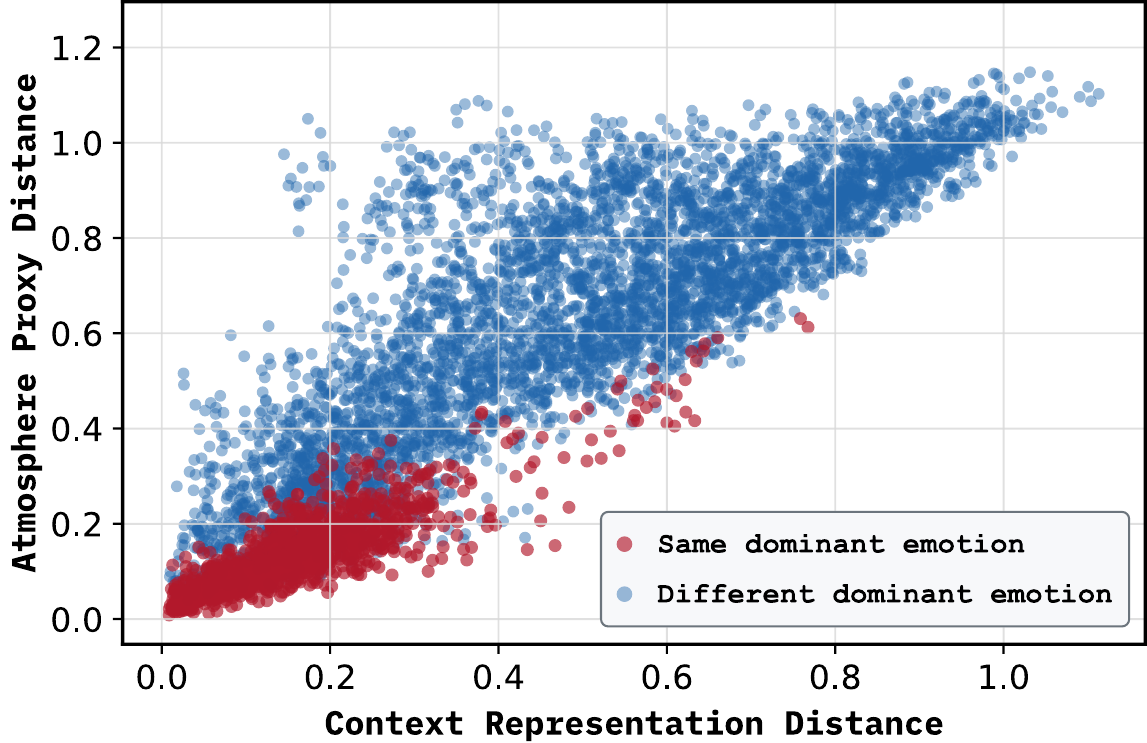}
    \caption{Visualization of global context versus affective atmosphere.
        Each point denotes a conversation pair.}
    \label{fig:gct_gea}
\end{figure}

The literature on social interaction describes atmosphere as an emergent affective quality of collective situations, shaped by context and turn-to-turn communication~\citep{jakobsadeh2025}.
Drawing on this perspective, atmosphere can broadly be defined as a latent, conversation-level emotional tendency shaped by conversational content and interactions among speakers.
As an affective property of conversation context, atmosphere characterizes the conversation's overall emotional orientation while allowing emotional differences between speakers and fluctuations across utterances~\cite{anderson2009a}.
Figure~\ref{fig:gct_gea} illustrates the correlation and difference between atmosphere and global context.
Generally, the differences in atmosphere are positively associated with differences in global context,  yet atmosphere exhibits relatively more stable characteristics even across substantially different contexts (more details can be found in Appendix~\ref{apx:global_vs_atmos}).

\begin{figure}[!t]
\centering
    \includegraphics[width=\columnwidth]{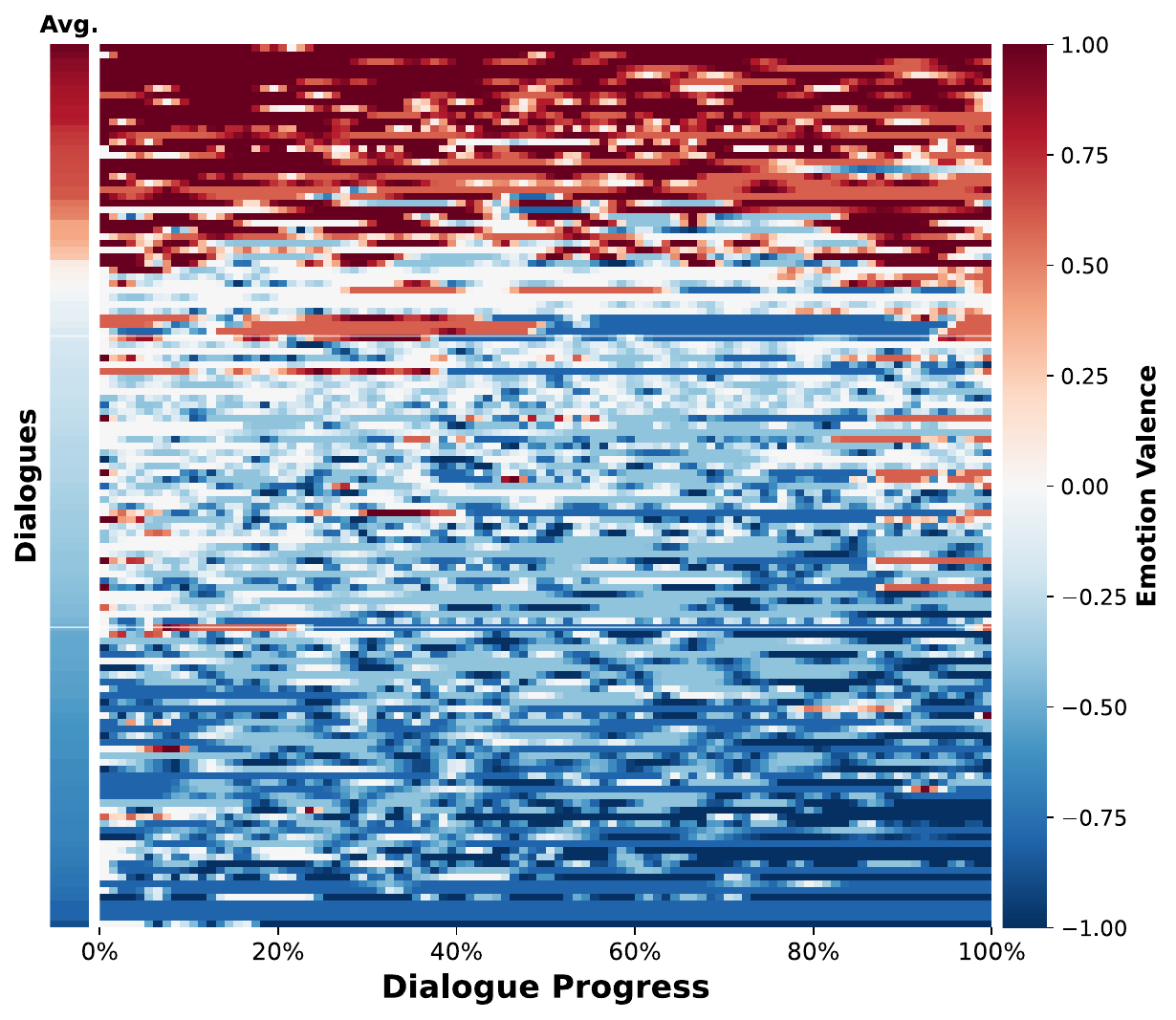}
    \caption{Emotion trajectories on IEMOCAP.
    Each row represents a complete conversation, with the horizontal axis showing normalized conversation progress.
    Colors denote utterance-level valence, from \textcolor{valenceblue}{negative} to \textcolor{valencered}{positive}, and the left strip shows the average valence of each conversation.}
    \label{fig:atmos_heatmap}
\end{figure}

Although atmosphere is latent within conversation context, utterance-level emotion trajectories can provide indirect observation for this underlying affective tendency.
Figure~\ref{fig:atmos_heatmap} visualizes valence trajectories on IEMOCAP, with each row representing a complete conversation and colors indicating utterance-level valence.
The trajectories show sustained positive or negative regions alongside local emotional fluctuations, suggesting that an overall affective tendency can persist despite changes in the emotions of individual utterances.
These observations naturally lead to two questions that this paper seeks to answer: \uline{Can affective atmosphere be explicitly estimated from conversation? What role does it play in conversational emotion understanding?}

The primary aim of this paper is to empirically investigate the extraction and use of affective atmosphere for ERC.
First, we propose \method, a graph-based atmosphere extractor that represents each conversation as a conversational graph over utterances and speaker interactions.
By filtering and fusing relation-specific graph signals, the extractor estimates a compact conversation-level affective atmosphere.
Second, to examine the role of atmosphere in conversational emotion understanding, we evaluate the estimated atmosphere in two downstream applications: guiding lightweight sequential emotion decoding and providing verbalized atmosphere cues for LLM-based ERC.
Finally, extensive experiments on three widely used ERC benchmarks show that the estimated atmosphere improves emotion prediction and supports more reliable judgments when local emotions deviate from the overall affective tendency.

The main contributions of this work are:
\begin{itemize}[leftmargin=*, itemsep=0pt, topsep=0pt, parsep=0pt, partopsep=0pt]
    \item We develop \method, a graph-based method that explicitly estimates affective atmosphere from conversational context and speaker interaction.
    \item We investigate two uses of the estimated atmosphere: initializing sequential emotion decoding and providing emotion cues for existing LLM-based ERC systems.
    \item Experiments on three ERC benchmarks show the benefits of atmosphere for conversational emotion prediction, particularly in cases of abrupt emotional shifts.
\end{itemize}
\section{Related Work}

Prior work on ERC has extensively modeled conversational context, emotion dynamics, and speaker interactions~\citep{poria2019a,wu2025a}.
Representative approaches include recurrent models for sequential context modeling~\citep{jiao2020,majumder2019,hu2021a}, Transformer-~\citep{zhong2019,li2020,mao2021} and PLM-based~\citep{kim2021a,wang2024a,yu2024} models for contextual representation learning, and graph-based models for explicit dependency modeling over conversational structures~\citep{chen2023,tu2024,ai2025,li2025a}.

Recent studies have increasingly explored LLM-based ERC~\citep{zhang2025a,fu2026,hassan2026,tu2026a}.
Representative directions include retrieval-augmented instruction tuning~\citep{lei2024,li2026a}, speaker-characteristic modeling~\citep{fu2025a}, and causal prompting~\citep{jing2026}.
TraceERC~\citep{xue2026} also explores atmosphere modeling through contextual exemplars.
These approaches enrich emotion prediction through task-specific training and additional contextual cues concerning speakers, emotion causes, and conversational affective tendencies.

Different from above, our work focuses on explicitly estimating conversation-level affective atmosphere and investigating its role in ERC.
Extended related work is provided in Appendix~\ref{apx:related_work}.

\begin{figure*}[!t]
    \centering
    \includegraphics[width=.85\textwidth]{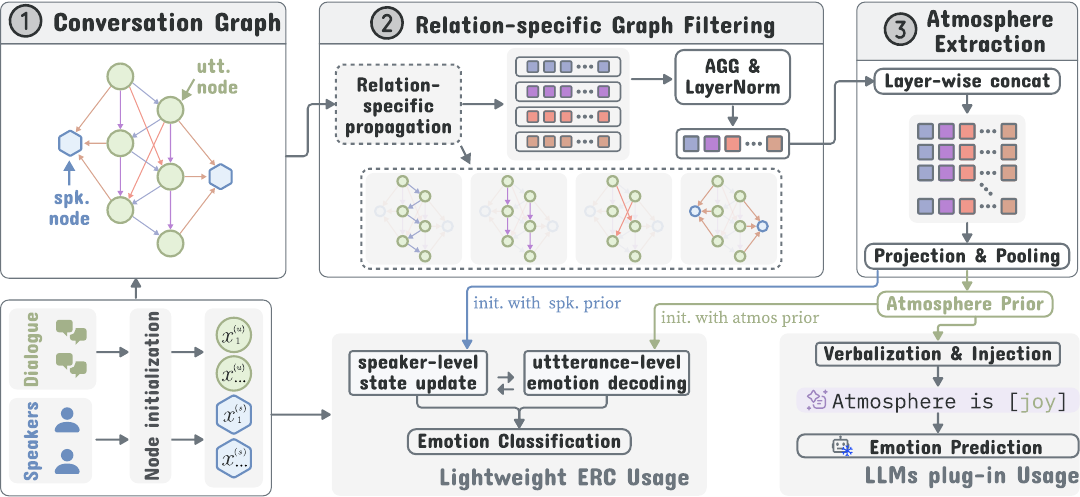}
    \caption{Overall architecture of our~\method.}
    \label{fig:archi}
\end{figure*}

\section{Methodology}\label{sec:method}

The proposed \method estimates a dialogue-level affective prior and uses it as shared conditioning information for utterance-level emotion prediction, as illustrated in Figure~\ref{fig:archi}.
Our methodology centers on two questions: how to estimate dialogue-level affective atmosphere and how to use it for emotion prediction.
We first formalize ERC and introduce atmosphere as a latent dialogue-level affective tendency (\S\ref{subsec:preliminary}).
We then develop an atmosphere-aware graph extractor that estimates this tendency from a conversational graph and encodes it as a compact affective prior (\S\ref{subsec:extractor}).
Finally, we use the estimated prior to guide lightweight ERC (\S\ref{subsec:lightweight}) and provide prompt-level plug-in cues for LLM-based ERC (\S\ref{subsec:plugin}).

\subsection{Preliminary}\label{subsec:preliminary}

\paragraph{Task Definition.}
A conversation is an ordered sequence of $N$ utterances $\mathcal{C} = (u_1, u_2, \dots, u_N)$. Each utterance $u_i$ is uttered by a speaker $s_{\pi(u_i)}$, where $\pi(\cdot)$ maps $u_i$ to the index of its corresponding speaker.
The conversation involves $M$ distinct speakers $(M \ge 2)$, denoted as $\{s_1, \dots, s_M\}$.
The goal of ERC is to assign an emotion label $y_i \in \mathcal{Y}$ to each utterance $u_i$, where $\mathcal{Y}$ is the set of predefined emotion categories.

\paragraph{Dialogue-level Affective Atmosphere.}
We treat atmosphere as a latent dialogue-level affective tendency within conversational context.
We represent its estimate as an embedding $\bm{a}\in\mathbb{R}^{d}$, where $d$ is the embedding dimension.
The atmosphere extractor is trained jointly with a lightweight ERC decoder under utterance-level emotion supervision.
The resulting embedding $\bm{a}$ serves as an affective prior for utterance-level emotion prediction.

\paragraph{Textual Feature Extraction.}
We encode each utterance $u_i \in \mathcal{C}$ into a feature embedding $\bm{x}_i^{(u)} \in \mathbb{R}^d$ using a pretrained RoBERTa~\citep{roberta}.
RoBERTa is used solely as a frozen feature extractor, with its parameters unchanged throughout the framework, thereby decoupling utterance-level feature learning from atmosphere modeling.

\subsection{Graph-based Atmosphere Extractor}\label{subsec:extractor}

To estimate dialogue-level affective atmosphere, we aggregate conversational evidence through a graph based on two considerations.
(1) \emph{Atmosphere is a dialogue-level signal} that should be estimated from the entire conversation, making graph modeling a natural choice.
(2) \emph{Atmosphere should emphasize stable affective tendencies}, while GNNs can act as low-pass filters over graph signals~\cite{gnn-low-pass-filter}, helping attenuate transient utterance-level variations.
Consequently, the atmosphere extractor consists of three steps, i.e., \circnum{1} conversational graph construction, \circnum{2} graph-based atmosphere aggregation, and \circnum{3} atmosphere extraction.

\paragraph{Conversational Graph Construction.} \label{subsubsec:graph}
We represent each conversation $\mathcal{C}$ as a heterogeneous graph $\mathcal{G}=(\mathcal{V}, \mathcal{E}, \mathcal{R})$, where $\mathcal{V}$, $\mathcal{E}$, and $\mathcal{R}$ denote the node set, edge set, and relation types, respectively.

\emph{Nodes.}
Following prior work~\citep{song2023}, we construct two disjoint node sets, $\mathcal{V}=\mathcal{V}^{(u)}\cup\mathcal{V}^{(s)}$, representing utterances and speakers.
Each utterance $u_i \in \mathcal{C}$ corresponds to an utterance node $v_i^{(u)}\in\mathcal{V}^{(u)}$, initialized with its textual representation $\bm{x}_i^{(u)}$.
Each speaker $s_j$ corresponds to a speaker node $v_j^{(s)}\in\mathcal{V}^{(s)}$, initialized with a learnable identity embedding $\bm{x}_j^{(s)} \in \mathbb{R}^d$.

\emph{Edges.}
Conversational dependencies are heterogeneous in nature, motivating multiple relation types $r \in \mathcal{R}$:
\begin{itemize}[leftmargin=2em, topsep=0pt, partopsep=0pt, itemsep=0pt, parsep=0pt]
    \item[$r_1$:] \textit{Inter-speaker dependency}  captures short-range cross-speaker influence. 
    We add a directed edge $\big(v_i^{(u)}, v_j^{(u)}\big)$ if $1 \le j-i \le W$ and $\pi(u_i) \neq \pi(u_j)$, where $W\in\mathbb{N}^+$ denotes the context window size.
    \item[$r_2$:] \textit{Intra-speaker dependency} models speaker-specific emotional continuity across consecutive speaker turns. 
    We add a directed edge $\big(v_i^{(u)}, v_j^{(u)}\big)$ if $u_j$ is the next utterance by the same speaker after $u_i$.
    \item[$r_3$:] \textit{Global semantic similarity} enables sparse long-range information exchange beyond local temporal neighborhoods. 
    We add an undirected edge $\big(v_i^{(u)}, v_j^{(u)}\big)$ if $i \neq j$ and $\cos \big(\bm{x}_i^{(u)}, \bm{x}_j^{(u)}\big)>\tau_s$, where $\tau_s$ is a semantic similarity threshold.
    \item[$r_4$:] \textit{Utterance-speaker affiliation} connects each utterance node with its corresponding speaker node to support speaker-aware affective aggregation. 
    We add a directed edge $\big(v_i^{(u)}, v_{\pi(u_i)}^{(s)}\big)$ for each utterance $u_i$.
\end{itemize}


\paragraph{Graph-based Atmosphere Aggregation.}\label{subsubsec:encoder}

To integrate atmosphere information while distinguishing different affective dependencies, each propagation layer first aggregates messages within individual relations and then combines the resulting representations.
For a target node $v_i \in \mathcal{V}$ and a relation $r \in \mathcal{R}$, let $\mathcal{N}_r(v_i)$ denote the set of neighbors connected to $v_i$ via relation $r$.
We update the relation-specific representation of $v_i$ as:
\begin{equation}
   \bm{h}_{i,r}^{(l)} = \mathrm{GNN}_{r}^{(l)}
    \big( \bm{h}_i^{(l-1)}, \{\bm{h}_j^{(l-1)}\}_{v_j \in \mathcal{N}_r(v_i)} \big),
\end{equation}
where $\mathrm{GNN}_r^{(l)}(\cdot)$ is a relation-specific GNN, such as GAT~\cite{gatv2}, that aggregates affective signals from neighbors under relation $r$, and $l\in\{1,\ldots,L\}$ denotes the GNN layer index.
The initial state of $v_i$ is given by $\bm{h}_i^{(0)}=\bm{x}_i$, while $\bm{h}_{j}^{(l-1)}$ aggregates the fused information from the layer before.
Different relations capture distinct affective dependencies and may contribute unequally to atmosphere estimation.
To preserve different affective dependencies, we fuse the representations $\{\bm{h}_{i,r}^{(l)} \mid r \in \mathcal{R}\}$ with an element-wise max aggregation:
\begin{equation}
    \bar{\bm{h}}_{i}^{(l)} = \mathrm{AGG}\big(\{\bm{h}_{i,r}^{(l)} \mid r \in \mathcal{R}\}\big).
\end{equation}
A non-linear activation and layer normalization then produce the updated node representation:
\begin{equation}
    \bm{h}_{i}^{(l)} =
    \mathrm{LayerNorm}\big(\mathrm{ReLU}(\bar{\bm{h}}_{i}^{(l)})\big).
\end{equation}
Notably, utterance nodes aggregate signals from $r_1$--$r_3$, whereas speaker nodes are updated exclusively via $r_4$.


\paragraph{Atmosphere Estimation.}

To estimate a stable dialogue-level atmosphere, we aggregate filtered representations across layers and project them into a compact embedding.
For $l\in\{0,\ldots,L\}$,  we denote the row-wise representation matrices of utterance and speaker nodes as $\bm{H}^{(u,l)} \in \mathbb{R}^{N \times d}$ and $\bm{H}^{(s,l)} \in \mathbb{R}^{M \times d}$, respectively, where $\bm{H}^{(u,l)}_{i,:} = (\bm{h}_{i}^{(u,l)})^\top$ and $\bm{H}^{(s,l)}_{j,:} = (\bm{h}_{j}^{(s,l)})^\top$.
We obtain compact graph-filtered representations by projecting the layer-wise concatenation of each node:
\begin{equation}
\begin{aligned}
    \bm{H}^{(u)}_{i,:} &= \mathrm{FC}_u
    \big( \bm{H}^{(u,0)}_{i,:} \Vert \cdots \Vert \bm{H}^{(u,L)}_{i,:} \big), \\
    \bm{H}^{(s)}_{j,:} &= \mathrm{FC}_s
    \big( \bm{H}^{(s,0)}_{j,:} \Vert \cdots \Vert \bm{H}^{(s,L)}_{j,:} \big),
\end{aligned}
\end{equation}
where $\bm{H}^{(u)} \in \mathbb{R}^{N \times d}$, $\bm{H}^{(s)} \in \mathbb{R}^{M \times d}$, $(\cdot\Vert\cdot)$ denotes feature-wise concatenation, and $\mathrm{FC}_u(\cdot)$ and $\mathrm{FC}_s(\cdot)$ are learnable linear projections.
We then summarize the contextualized utterance representations into a dialogue-level atmosphere representation:
\begin{equation}
    \bm{a} = \mathrm{Pool}(\bm{H}^{(u)}).
\end{equation}
where $\mathrm{Pool}(\cdot)$ performs element-wise max pooling over the $N$ utterance nodes.
The resulting vector $\bm{a}\in\mathbb{R}^d$ serves as the estimated atmosphere prior for downstream emotion prediction.
We separately retain $\bm{H}^{(s)}$ as auxiliary speaker-conditioned priors for lightweight ERC, with each row providing the initial representation for its corresponding speaker.

\subsection{Atmosphere-guided Lightweight ERC}\label{subsec:lightweight}

Based on the extracted atmosphere prior, we instantiate a lightweight ERC model to examine whether the atmosphere can directly benefit conversational emotion prediction.
The lightweight model uses the extracted priors as initialization signals for dynamic emotion decoding.
Specifically, the atmosphere embedding $\bm{a}$ initializes the utterance-level temporal decoder, while the speaker-conditioned priors $\bm{H}^{(s)}$ initialize speaker-specific states.
The original utterance representations $\{\bm{x}^{(u)}_i\}_{i=1}^{N}$ remain the inputs to the decoder, ensuring that local emotion dynamics are modeled from utterance semantics while being guided by dialogue-level and speaker-conditioned affective priors.

\paragraph{Atmosphere-conditioned Emotion Dynamics.}
We design an atmosphere-conditioned decoder to model utterance-level emotion dynamics under the dialogue-level affective prior.
For each utterance $u_i$, the decoder maintains a latent state $\bm{o}^{(u)}_i$ that captures the evolving emotional trajectory of the conversation.
At each turn, the state is updated from the utterance representation $\bm{x}^{(u)}_i$, while the initial decoding state is determined by the estimated atmosphere.
Formally, we use a simple BiGRU:
\begin{equation}
    \bm{o}_i^{(u)}
    = \mathrm{BiGRU}_u\big(\bm{x}_i^{(u)}; \bm{a}\big).
\end{equation}
In practice, $\bm{a}$ is used to initialize both the forward and backward hidden states of the BiGRU, so that local utterance-level dynamics are decoded under a dialogue-level affective prior.

\paragraph{Speaker-specific State Dynamics.}
We further introduce a speaker-state decoder to track speaker-conditioned affective dynamics throughout the conversation.
Each row $\bm{h}^{(s)}_j$ of $\bm{H}^{(s)}$ initializes the state of speaker $s_j$.
During decoding, a speaker state is updated only when that speaker produces an utterance.
At turn $i$, given the speaker index $j=\pi(u_i)$ and the utterance-level state $\bm{o}^{(u)}_i$, we update the speaker state as:
\begin{equation}
    \bm{h}^{(s)}_{j,i} = \mathrm{GRU}_s \big( \bm{o}^{(u)}_i, \bm{h}^{(s)}_{j,i-1} \big),
\end{equation}
while keeping the states of all other speakers unchanged.

\paragraph{Atmosphere-guided Emotion Prediction.}
For each utterance $u_i$, emotion prediction is based on both the atmosphere-conditioned state $\bm{o}^{(u)}_i$ and the updated speaker state $\bm{h}^{(s)}_{\pi(u_i)}$:
\begin{equation}
    \bm{o}_i = \big[ \bm{o}^{(u)}_i; \bm{h}^{(s)}_{\pi(u_i)} \big].
\end{equation}
The fused representation $\bm{o}_i$ is fed into a linear classifier followed by softmax to obtain the emotion label $\hat{y}\in\mathcal{Y}$.
The lightweight model is trained with the standard cross-entropy loss over all utterances.

\subsection{Atmosphere Plug-in for LLM-based ERC}\label{subsec:plugin}

We further examine whether the extracted atmosphere can serve as a prompt-level plug-in for LLM-based ERC methods.
Since prompt-based LLMs cannot directly consume the continuous vector $\bm{a}$, we verbalize it into a discrete descriptor and inject it into the original ERC prompt without modifying the LLM backbone or decoding procedure.

\paragraph{Atmosphere Verbalization.}

To make the continuous atmosphere vector usable by LLM-based methods, we discretize it into a textual atmosphere descriptor.
Since atmosphere has no direct annotation, we first derive a dialogue-level proxy target from the dominant emotion in each training dialogue:
\begin{equation}
    z^\star = \arg\max_{y \in \mathcal{Y}} \sum_{i=1}^{N} \mathbb{I}(y_i = y).
\end{equation}
We then train a lightweight verbalization head to predict this proxy from the atmosphere vector $\bm{a}$
\begin{equation}
\begin{aligned}
    \bm{q} &= \mathrm{softmax}\bigl(\mathrm{FC}_{v}(\bm{a})\bigr), \\
    \widehat{z} &= \arg\max_{y \in \mathcal{Y}} \bm{q}[y].
\end{aligned}
\end{equation}
The head is optimized to predict $z^\star$ from $\bm{a}$ on the training set, and $\widehat{z}$ is used as the textual descriptor for prompt injection.
During inference, $\widehat{z}$ is computed solely from $\bm{a}$, which is extracted from the input dialogue without access to utterance-level emotion labels.

\definecolor{promptbg}{HTML}{E6F0FA}
\newtcolorbox{promptblock}{
    colback=promptbg,
    colframe=promptbg,
    boxrule=0pt,
    arc=2pt,
    left=4pt,
    right=4pt,
    top=2pt,
    bottom=2pt,
    boxsep=0pt,
    before skip=2pt,
    after skip=2pt,
    fontupper=\small\ttfamily,
    enhanced,
    breakable
}

\paragraph{Prompt-level Atmosphere Injection.}
Given a target utterance $u_i$ and its dialogue context, we inject the predicted descriptor $\widehat{z}$ into the original LLM-based ERC prompt as an explicit atmosphere prior.
Specifically, we prepend a short atmosphere instruction to the original prompt while keeping the remaining prompt format unchanged:
\begin{promptblock}
The overall affective atmosphere of this dialogue is \texttt{<ATMOSPHERE>}.
\end{promptblock}
\noindent Here, \texttt{<ATMOSPHERE>} is filled with the predicted descriptor $\widehat{z}$.
This plug-in setting allows us to evaluate whether the atmosphere prior provides complementary information under the same backbone, prompt structure, and decoding configuration.
Detailed prompt templates are provided in Appendix~\ref{apx:prompt}.

\section{Experimental Setups}

\subsection{Datasets}

We evaluate our model on three widely used benchmark datasets for ERC, namely \textbf{IEMOCAP}~\citep{busso2008}, \textbf{MELD}~\citep{poria2019}, and \textbf{EmoryNLP}~\citep{zahiri2017}.
Dataset statistics are summarized in Table~\ref{tab:dataset}, and detailed dataset descriptions are provided in Appendix~\ref{app:dataset}.

\begin{table}[!t]
    \centering
    \scriptsize
    \setlength{\tabcolsep}{2pt}
    \begin{tabular*}{\columnwidth}{@{\extracolsep{\fill}}lcccc@{}}
    \toprule
    \multirow{2}{*}{\textbf{Dataset}} 
    & \textbf{\# Conversations} & \textbf{\# Utterances} & \textbf{Avg.} & \textbf{Avg.} \\
    & \textbf{\colorbox{gray!15}{\makebox[2em]{train}} / 
    \colorbox{gray!25}{\makebox[1.5em]{val}} / 
    \colorbox{gray!35}{\makebox[1.5em]{test}}} 
    & \textbf{\colorbox{gray!15}{\makebox[2em]{train}} / 
    \colorbox{gray!25}{\makebox[1.5em]{val}} / 
    \colorbox{gray!35}{\makebox[1.5em]{test}}} 
    & \textbf{Turns} & \textbf{Spk.} \\
    \midrule
    IEMOCAP     
    & \colorbox{gray!15}{\makebox[2em]{120}} / 
      \colorbox{gray!25}{\makebox[1.5em]{\vphantom{0}--}} / 
      \colorbox{gray!35}{\makebox[1.5em]{31}}
    & \colorbox{gray!15}{\makebox[2em]{5810}} / 
      \colorbox{gray!25}{\makebox[1.5em]{\vphantom{0}--}} / 
      \colorbox{gray!35}{\makebox[1.5em]{1623}}
    & 49.2 & 2 \\

    MELD        
    & \colorbox{gray!15}{\makebox[2em]{1039}} / 
      \colorbox{gray!25}{\makebox[1.5em]{114}} / 
      \colorbox{gray!35}{\makebox[1.5em]{280}}
    & \colorbox{gray!15}{\makebox[2em]{9989}} / 
      \colorbox{gray!25}{\makebox[1.5em]{1109}} / 
      \colorbox{gray!35}{\makebox[1.5em]{2610}}
    & 9.6  & 2.7 \\

    EmoryNLP    
    & \colorbox{gray!15}{\makebox[2em]{713}} / 
      \colorbox{gray!25}{\makebox[1.5em]{99}} / 
      \colorbox{gray!35}{\makebox[1.5em]{85}}
    & \colorbox{gray!15}{\makebox[2em]{9934}} / 
      \colorbox{gray!25}{\makebox[1.5em]{1344}} / 
      \colorbox{gray!35}{\makebox[1.5em]{1328}}
    & 14.1 & 3.2 \\

    \bottomrule
    \end{tabular*}
    \caption{Dataset statistics.}
    \label{tab:dataset}
\end{table}

\subsection{Baselines}


We organize the compared methods according to the two evaluation settings.

For the lightweight ERC setting, we compare \method with representative non-LLM ERC baselines covering five methodological groups: sequence-based methods, including SGED~\citep{bao2022} and SACL-LSTM~\citep{hu2023};
graph-based methods, including DAG-ERC~\citep{shen2021}, SUNET~\citep{song2023}, and FrameERC~\citep{li2025d}; knowledge- and cause-aware methods, including CauAIN~\citep{zhao2022}, SKIER~\citep{li2023}, and LECM~\citep{lu2025b}; attention-based multimodal fusion methods, including CFN-ESA~\citep{li2024c} and EMART~\citep{seong2026}; and label-aware learning methods, including EACL~\citep{yu2024} and SEL+SACL~\citep{lai2026}.

For the LLM-based plug-in setting, we evaluate whether the estimated atmosphere improves performance in InstructERC~\citep{lei2024}, GraphEMO~\citep{tu2026a}, Causal-ERC~\citep{jing2026}, and TraceERC~\cite{xue2026}.
Detailed descriptions of the baselines are provided in Appendix~\ref{app:baselines}.

\subsection{Experiment Details}

We use only the textual modality and report weighted-average F1 scores on all datasets.
We denote the lightweight ERC model as AtmosERC and the LLM plug-in variant as AtmosERC-P. 
Since these datasets lack direct atmosphere annotations, we use each dialogue’s dominant emotion as a proxy for atmosphere verbalization and diagnostic analyses. 
This proxy is not used to supervise the atmosphere extractor. 
For lightweight ERC, we average results over multiple random seeds.
For LLM plug-in evaluation, we keep the backbone, base prompt, and decoding settings unchanged, with the predicted atmosphere descriptor as the only additional input. 
Detailed settings are provided in Appendix~\ref{app:hyper}.

\section{Results and Analysis}

We structure our evaluation around three research questions:
\textbf{RQ1}: To what extent does atmosphere guidance improve lightweight ERC?
\textbf{RQ2}: How effective is the estimated atmosphere as a plug-in prior for LLM-based ERC?
\textbf{RQ3}: Under what affective conditions does atmosphere help or hinder emotion recognition?

\begin{table}[!t]
\centering
\footnotesize
\setlength{\tabcolsep}{2pt}
\begin{tabularx}{\columnwidth}{XYYY}
    \toprule
    \textbf{Method} & \textbf{IEMOCAP} & \textbf{MELD} & \textbf{EmoryNLP} \\
    \midrule
    SGED & 68.53 & 65.46 & 40.24 \\
    SACL-LSTM & 69.22 & 66.45 & 39.65 \\
    DAG-ERC & 68.03 & 63.65 & 39.02 \\
    SUNET & 68.96 & 64.03 & 39.89 \\
    FrameERC & 70.67 & 66.33 & -- \\
    CauAIN & 67.61 & 65.46 & -- \\
    SKIER & 68.42 & \uline{67.39} & 40.07 \\
    LECM & 68.52 & 65.57 & --\\
    CFN-ESA & \uline{71.04} & 66.70 & -- \\
    EMART & 70.93 & 64.19 & -- \\
    EACL & 70.41 & 67.12 & 40.24 \\
    SEL+SACL & 69.51 & 66.52 & \uline{40.57} \\
    \midrule
    \method             & \first{71.29} & \first{69.22} & \first{40.75} \\
    \bottomrule
\end{tabularx}
\caption{Overall performance of non-LLM ERC methods on three benchmark datasets. {\setlength{\fboxsep}{2pt}\colorbox{1st}{\textbf{Shading}}} and \uline{underlining} indicate the best and second-best results.}
\label{tab:main_result}
\end{table}

\subsection{Atmosphere Improves Lightweight ERC}

Table~\ref{tab:main_result} reports the overall performance of \method against representative non-LLM ERC baselines.
\method achieves the highest performance among the baselines on all three datasets.
Specifically, our \method outperforms the strongest graph-based baseline on each dataset by $+0.62$, $+2.89$, and $+0.86$ points on IEMOCAP, MELD, and EmoryNLP, respectively, compared with FrameERC on the first two datasets and SUNET on the third.
\method also compares favorably with approaches that fine-tune PLMs or incorporate external knowledge.
Compared with EACL, which fine-tunes a pretrained language model, \method improves weighted-F1 by $+0.88$, $+2.10$, and $+0.51$ points across the three datasets.
It also surpasses the knowledge-enhanced SKIER by $+2.87$, $+1.83$, and $+0.68$ points, respectively.
Overall, these results highlight the effectiveness of using conversational graphs to provide an explicit dialogue-level atmosphere prior for ERC.

\begin{table}[!t]
\centering
\footnotesize
\setlength{\tabcolsep}{2pt}
\begin{tabular*}{\columnwidth}{l@{\extracolsep{\fill}}lll}
    \toprule
    \textbf{Method} & \textbf{IEMOCAP} & \textbf{MELD} & \textbf{EmoryNLP} \\
    \midrule
    \rowcolor{gray!25}
    \multicolumn{4}{c}{\textit{General-purpose LLMs}} \\
    \midrule
    GPT-5.4-mini & 57.72 & 62.84 & 37.61 \\
    ~+ \method-P
        & 60.12 \gain{480}{+2.40}
        & 63.72 \gain{176}{+0.88}
        & 38.36 \gain{150}{+0.75} \\
    Gemini-3-flash & 59.82 & 63.94 & 40.87 \\
    ~+ \method-P
        & 61.22 \gain{280}{+1.40}
        & 64.56 \gain{124}{+0.62}
        & 40.98 \gain{22}{+0.11} \\
    DeepSeek-v4.1-flash & 56.99 & 67.18 & 38.05 \\
    ~+ \method-P
        & 60.01 \gain{604}{+3.02}
        & 67.95 \gain{154}{+0.77}
        & 38.90 \gain{170}{+0.85} \\
    \midrule
    \rowcolor{gray!25}
    \multicolumn{4}{c}{\textit{ERC-specific LLM Methods}} \\
    \midrule
    InstructERC & 71.39 & 69.15 & 41.37 \\
    ~+ \method-P
        & 71.84 \gain{90}{+0.45}
        & 69.78 \gain{126}{+0.63}
        & 41.76 \gain{78}{+0.39} \\
    GraphEMO & 70.19 & 68.27 & 38.92 \\
    ~+ \method-P
        & 71.13 \gain{188}{+0.94}
        & 68.67 \gain{80}{+0.40}
        & 39.29 \gain{74}{+0.37} \\
    Causal-ERC & 69.13 & 68.10 & 40.64 \\
    ~+ \method-P
        & 71.83 \gain{540}{+2.70}
        & 69.18 \gain{216}{+1.08}
        & 40.94 \gain{60}{+0.30} \\
    TraceERC & 69.41 & 69.04 & 41.34 \\
    ~+ \method-P
        & 70.59 \gain{236}{+1.18}
        & 69.88 \gain{168}{+0.84}
        & 41.41 \gain{14}{+0.07} \\
    \bottomrule
\end{tabular*}
\caption{Performance with and without the prompt-level atmosphere plug-in.}
\label{tab:llm_plugin}
\end{table}

\subsection{Atmosphere Enhances LLM-based ERC}

For a fair paired comparison, we evaluate each LLM backbone and its atmosphere-augmented variant with the same base prompts and decoding parameters, with the descriptor injection as the sole differentiator.
For ERC-specific LLMs, we further re-implement their native pipelines and apply the same paired evaluation protocol.
As shown in Table~\ref{tab:llm_plugin}, \method-P consistently improves all evaluated general-purpose LLMs and ERC-specific LLM methods across IEMOCAP, MELD, and EmoryNLP.
The gains are generally more pronounced on IEMOCAP and MELD, reaching up to $+3.02$ and $+1.08$, respectively.
There were no significant improvements on EmoryNLP, which may be explained by the fact that the atmosphere extraction is insufficient, as suggested by the lower agreement between predicted atmosphere descriptors and dominant-emotion proxies reported in Appendix~\ref{app:proxy_atmosphere}.
In summary, these results show that dialogue-level affective atmosphere provides useful global guidance for ERC, benefiting both general-purpose backbones and methods already tailored to the task.
Appendix~\ref{app:proxy_atmosphere} further provides a proxy-based diagnostic analysis, indicating that stronger atmosphere estimation could further improve LLM performance for ERC.

\subsection{Atmosphere Benefits Are Conditional}


\paragraph{Dominant Emotion Share.}

\begin{table}[!t]
\centering
\footnotesize
\setlength{\tabcolsep}{2pt}
\begin{tabular*}{\columnwidth}{l@{\extracolsep{\fill}}ccc}
    \toprule
    \textbf{Strength} & \textbf{IEMOCAP} & \textbf{MELD} & \textbf{EmoryNLP} \\
    \midrule
    Low    & +0.62 \textcolor{1st!62}{$\blacktriangle$} & +7.11 \textcolor{1st!711}{$\blacktriangle$} & -0.61 \textcolor{myred!61}{$\blacktriangledown$} \\
    Medium & +3.40 \textcolor{1st!340}{$\blacktriangle$} & +5.92 \textcolor{1st!592}{$\blacktriangle$} & +0.62 \textcolor{1st!62}{$\blacktriangle$} \\
    High   & +5.40 \textcolor{1st!540}{$\blacktriangle$} & +1.19 \textcolor{1st!119}{$\blacktriangle$} & +1.23 \textcolor{1st!123}{$\blacktriangle$} \\
    \bottomrule
\end{tabular*}
\caption{Weighted-F1 gains of the atmosphere prior over the context prior $\bm{c}$ across dominant-emotion concentration groups.}
\label{tab:strength_split}
\end{table}

We first examine whether atmosphere provides larger ERC gains in dialogues with a higher dominant emotion share.
We quantify this share as the proportion of utterances labeled with the dialogue's most frequent gold emotion, using it as a proxy for atmosphere concentration.
Within each dataset, we partition the test dialogues into three equally sized groups with low, medium, and high shares.
We conduct this analysis using lightweight AtmosERC, with a variant conditioned on the original context representation $\bm{c}$ serving as the baseline without explicit atmosphere modeling.
Table~\ref{tab:strength_split} reports weighted-F1 gains over this baseline.
As the dominant emotion share increases, gains rise from $+0.62$ to $+5.40$ on IEMOCAP and from $-0.61$ to $+1.23$ on EmoryNLP.
Surprisingly, the gains decrease from $+7.11$ to $+1.19$ on MELD, which may be because MELD's short dialogues and brief utterances provide limited evidence for estimating a dialogue-level atmosphere.
These results suggest that atmosphere benefits extend beyond dialogues with a highly dominant emotion and that a higher dominant emotion share does not uniformly correspond to larger gains.

\begin{table}[!t]
\centering
\footnotesize
\setlength{\tabcolsep}{2pt}
\begin{tabular*}{\columnwidth}{l@{\extracolsep{\fill}}lll}
    \toprule
    \textbf{Methods} & \textbf{IEMOCAP} & \textbf{MELD} & \textbf{EmoryNLP} \\
    \midrule
    \rowcolor{gray!25}
    \multicolumn{4}{c}{\textit{Correct prediction on} $u_{i+1}$} \\
    \midrule
    DeepSeek-v4.1-flash & 65.84 & 68.40 & 44.91 \\
    \quad +\method-P
        & 67.70 \gain{372}{+1.86}
        & 72.00 \gain{720}{+3.60}
        & 50.30 \gain{1078}{+5.39} \\
    Causal-ERC & 74.53 & 72.80 & 41.32 \\
    \quad +\method-P
        & 76.40 \gain{374}{+1.87}
        & 74.40 \gain{320}{+1.60}
        & 41.92 \gain{120}{+0.60} \\
    \midrule
    \rowcolor{gray!25}
    \multicolumn{4}{c}{\textit{Correct prediction on} $(u_i,u_{i+1})$} \\
    \midrule
    DeepSeek-v4.1-flash & 37.32 & 39.60 & 16.17 \\
    \quad +\method-P
        & 37.32 \gain{0}{+0.00}
        & 40.00 \gain{80}{+0.40}
        & 17.96 \gain{358}{+1.79} \\
    Causal-ERC & 44.10 & 42.40 & 11.38 \\
    \quad +\method-P
        & 43.48 \loss{6.2}{-0.62}
        & 41.60 \loss{8}{-0.80}
        & 11.98 \gain{120}{+0.60} \\
    \midrule
    \rowcolor{gray!25}
    \multicolumn{4}{c}{\textit{Correct prediction on} $(u_{i-1},u_i,u_{i+1})$} \\
    \midrule
    DeepSeek-v4.1-flash & 29.19 & 27.60 & 11.98 \\
    \quad +\method-P
        & 29.19 \gain{0}{+0.00}
        & 30.00 \gain{480}{+2.40}
        & 14.97 \gain{598}{+2.99} \\
    Causal-ERC & 40.37 & 32.00 & ~~7.78 \\
    \quad +\method-P
        & 37.27 \loss{31}{-3.10}
        & 34.00 \gain{400}{+2.00}
        & ~~8.98 \gain{240}{+1.20} \\
    \bottomrule
\end{tabular*}
\caption{Accuracy (\%) on triplets containing a local emotional deviation.}
\label{tab:transition_analysis}
\end{table}

\paragraph{Recovery after Local Deviations.}

We then examine whether the extracted atmosphere helps recognize returns to the dominant emotion after local emotional deviations.
For each dialogue, we evaluate LLM predictions on consecutive utterance triplets $(u_{i-1},u_i,u_{i+1})$ satisfying $y_{i-1}=y_{i+1}=y^\ast$ and $y_i\neq y^\ast$, where $y^\ast$ is the dialogue's dominant emotion and $u_i$ represents a local emotional deviation.
As shown in Table~\ref{tab:transition_analysis}, \method-P improves final-utterance accuracy across all three datasets for both models, with gains ranging from $1.86$ to $5.39$ percentage points for DeepSeek-v4.1-flash and from $0.60$ to $1.87$ points for Causal-ERC.
Changes in pair- and triplet-level accuracy are mixed: DeepSeek-v4.1-flash's full-triplet accuracy increases by $2.40$ points on MELD, whereas Causal-ERC's full-triplet accuracy decreases by $3.10$ points on IEMOCAP.
These results suggest that atmosphere can support recognition of returns to the dominant emotion, while improvements in jointly recognizing deviations and returns are not consistent across models and datasets.

\begin{table}[!t]
\centering
\footnotesize
\setlength{\tabcolsep}{2pt}
\begin{tabular*}{\columnwidth}{l@{\extracolsep{\fill}}lll}
    \toprule
    \textbf{Methods} & \textbf{IEMOCAP} & \textbf{MELD} & \textbf{EmoryNLP} \\
    \midrule
    \rowcolor{gray!25}
    \multicolumn{4}{c}{\textit{Correct prediction on} $u_{i+1}$} \\
    \midrule
    DeepSeek-v4.1-flash & 56.56 & 64.86 & 39.05 \\
    \quad +\method-P
        & 55.74 \loss{8.2}{-0.82}
        & 64.86 \gain{0}{+0.00}
        & 38.41 \loss{6.4}{-0.64} \\
    Causal-ERC & 59.84 & 64.46 & 40.63 \\
    \quad +\method-P
        & 60.66 \gain{164}{+0.82}
        & 66.67 \gain{442}{+2.21}
        & 40.32 \loss{3.1}{-0.31} \\
    \midrule
    \rowcolor{gray!25}
    \multicolumn{4}{c}{\textit{Correct prediction on} $(u_i,u_{i+1})$} \\
    \midrule
    DeepSeek-v4.1-flash & 37.70 & 42.17 & 24.13 \\
    \quad +\method-P
        & 39.34 \gain{328}{+1.64}
        & 43.17 \gain{200}{+1.00}
        & 24.76 \gain{126}{+0.63} \\
    Causal-ERC & 50.00 & 47.39 & 23.17 \\
    \quad +\method-P
        & 50.82 \gain{164}{+0.82}
        & 48.39 \gain{200}{+1.00}
        & 24.13 \gain{192}{+0.96} \\
    \midrule
    \rowcolor{gray!25}
    \multicolumn{4}{c}{\textit{Correct prediction on} $(u_{i-1},u_i,u_{i+1})$} \\
    \midrule
    DeepSeek-v4.1-flash & 18.85 & 27.31 & ~~8.89 \\
    \quad +\method-P
        & 18.85 \gain{0}{+0.00}
        & 25.10 \loss{22.1}{-2.21}
        & ~~8.57 \loss{3.2}{-0.32} \\
    Causal-ERC & 24.59 & 29.92 & ~~8.89 \\
    \quad +\method-P
        & 22.95 \loss{16.4}{-1.64}
        & 29.32 \loss{6}{-0.60}
        & ~~8.89 \gain{0}{+0.00} \\
    \bottomrule
\end{tabular*}
\caption{Accuracy (\%) on triplets with persistent local emotional shifts.}
\label{tab:shift_interference}
\end{table}

\paragraph{Interference under Persistent Shifts.}
In contrast to the previous experiment, we examine whether the estimated atmosphere prior $\bm{a}$ biases subsequent predictions toward the dominant emotion when a local emotional shift persists.
Using the same paired LLM setting, we evaluate consecutive utterance triplets $(u_{i-1},u_i,u_{i+1})$ satisfying $y_{i-1}=y^\ast$, $y_i\neq y^\ast$, and $y_{i+1}\neq y^\ast$.
As shown in Table~\ref{tab:shift_interference}, changes in final-utterance accuracy range from $-0.82$ to $+2.21$ percentage points, in contrast to the consistent improvements observed for ABA triplets.
Nevertheless, \method-P improves joint accuracy on $(u_i,u_{i+1})$ for both models across all three datasets, with gains ranging from $0.63$ to $1.64$ points.
These results suggest that atmosphere can support joint recognition of consecutive non-dominant emotions, although this benefit does not consistently extend to the complete triplet.

\begin{table}[!t]
\centering
\footnotesize
\setlength{\tabcolsep}{2pt}
\begin{tabular*}{\columnwidth}{@{\extracolsep{\fill}}lccc}
    \toprule
    \textbf{Variants} & \textbf{IEMOCAP} & \textbf{MELD} & \textbf{EmoryNLP} \\
    \midrule
    \rowcolor{gray!25}
    \multicolumn{4}{c}{\textit{Atmosphere prior and decoding}} \\
    \midrule
    Full model  & \textbf{71.29} & \textbf{69.22} & \textbf{40.75} \\
    w/o atmosphere prior    & 67.86 & 63.63 & 40.12 \\
    w/o speaker prior   & 67.27 & 63.62 & 39.12 \\
    \midrule
    \rowcolor{gray!25}
    \multicolumn{4}{c}{\textit{Graph structure}} \\
    \midrule
    w/o $r_1$      & 67.94 & 63.44 & 39.07 \\
    w/o $r_2$      & 67.99 & 63.54 & 39.32 \\
    w/o $r_3$      & 69.07 & 63.49 & 38.96 \\
    w/o $(r_1~\&~r_2)$ & 67.92 & 63.57 & 38.97 \\
    w/o $(r_1~\&~r_3)$ & 68.10 & 63.45 & 38.95 \\
    w/o $(r_2~\&~r_3)$ & 68.35 & 63.49 & 38.80 \\
    \bottomrule
\end{tabular*}
\caption{Ablation results.}
\label{tab:abla}
\end{table}

\subsection{Ablation of Atmosphere Modeling}
Table~\ref{tab:abla} evaluates atmosphere priors, speaker-specific decoding, and graph structure.
Replacing the learned atmosphere prior $\bm{a}$ with a context vector $\bm{c}$ pooled from initial utterance representations reduces weighted-F1 by $3.22$ points on average.
Removing the speaker-specific GRU yields an average drop of $3.75$ points.

For atmosphere estimation, we ablate utterance--utterance relations $r_1$--$r_3$ while retaining the utterance--speaker relation $r_4$.
Removing $r_1$, $r_2$, or $r_3$ yields average drops of $3.60$, $3.47$, and $3.25$ points, respectively; removing $(r_1,r_2)$, $(r_1,r_3)$, or $(r_2,r_3)$ yields drops of $3.60$, $3.59$, and $3.54$ points.
All variants underperform the full model across the three datasets, supporting the contributions of atmosphere priors, speaker-specific state modeling, and heterogeneous conversational relations to ERC.

\section{Conclusion}

In this work, we investigate the explicit estimation of conversation-level affective atmosphere and its role in ERC.
We propose \method, a graph-based extractor that explicitly estimates conversation-level affective atmosphere.
We investigate its role in ERC through lightweight sequential decoder initialization and predicted dominant-emotion cues for LLM-based ERC.
Experiments on three benchmarks show improved emotion prediction across both applications and more reliable predictions when local emotions deviate from the overall affective tendency.
These findings suggest that atmosphere can serve as a reusable affective prior for conversational emotion understanding.


\bibliography{bib/custom,bib/erc,bib/sa,bib/other}

\appendix
\makeatletter
\def\input@path{{appendix/}}
\makeatother
\section{Extended Related Work}\label{apx:related_work}

\paragraph{Context Modeling in Conversation.}
Conversations contain complex relational structures arising from dialogue history, speaker interactions, and long-range dependencies~\citep{poria2019a}.
Early ERC methods mainly modeled contextual dependencies with recurrent architectures~\citep{hazarika2018,majumder2019}, while later approaches adopted Transformers and pretrained language models to aggregate broader conversational context~\citep{zhong2019,li2020,kim2021a}.
More recently, graph-based methods explicitly construct conversational graphs over utterances or speakers, using relational edges or graph spectral operations to model structural dependencies~\citep{ghosal2019,shen2021,song2023,li2025d}.
Other approaches incorporate commonsense knowledge and emotion-cause information~\citep{zhao2022,li2023,lu2025b}, while multimodal fusion, contrastive learning, and emotion-label augmentation further enhance emotion representations~\citep{li2024c,seong2026,yu2024,lai2026}.
Although these methods have substantially advanced contextual modeling for ERC, they usually use context to refine utterance-level representations.
The stable affective component of dialogue-level context is rarely separated from local utterance features or formulated as a reusable global prior.
Our work differs by explicitly defining dialogue-level affective atmosphere and estimating it as an affective prior for downstream emotion prediction.

\paragraph{Emotional Dynamics Modeling.}
Another line of work focuses on utterance-level emotional dynamics, where intra-speaker emotional continuity and inter-speaker influence jointly shape emotion evolution~\citep{poria2019a,ghosal2019}.
Existing methods track such dynamics through recurrent networks, attention mechanisms, speaker-aware modeling, or personality-related attributes~\citep{hazarika2018,majumder2019,bao2022,hu2023,wang2024a,li2024c}.
Graph-based models further introduce speaker-related signals by using speaker embeddings, speaker nodes, or relation-specific edges~\citep{ghosal2019,song2023}.
However, emotional dynamics are not purely local.
A dialogue may maintain a relatively stable affective tendency while individual utterances fluctuate due to transient contextual triggers.
Existing dynamic modeling methods mainly focus on how emotions evolve across turns, but rarely distinguish local emotional fluctuations from the dialogue-level affective atmosphere underlying the conversation.
In contrast, our method uses the estimated atmosphere as a dialogue-level prior to guide local emotion decoding.

\paragraph{LLM-based ERC.}
Recent studies have increasingly explored LLM-based ERC~\citep{lei2024,tu2026a,jing2026,xue2026}.
InstructERC~\citep{lei2024} combines instruction tuning with retrieval augmentation and auxiliary tasks for conversational emotion prediction.
GraphEMO~\citep{tu2026a} unifies multimodal emotion recognition, emotion-cause reasoning, and emotional support through hierarchical task-specific priors and dual graph attention.
Causal-ERC~\citep{jing2026} integrates multimodal contextual representations and introduces causal prompts tailored to utterance-level causal types.
TraceERC~\citep{xue2026} incorporates speaker personality and predicted emotional context as contextual exemplars, providing atmosphere-related guidance for conversational emotion recognition.
These methods illustrate how task-specific training and explicit auxiliary information can support LLM-based emotion prediction.

\paragraph{Affective Atmosphere.}
Dialogue-level affective atmosphere is related to, but distinct from, conventional context modeling and emotional dynamics modeling.
Context modeling focuses on using surrounding utterances, speakers, or external knowledge to support utterance-level prediction~\citep{ghosal2019,shen2021,song2023,li2023,li2025d}, while emotional dynamics modeling focuses on how emotions change over time and across speakers~\citep{hazarika2018,majumder2019,bao2022,hu2023,li2024c}.
Research on affective atmospheres and social interaction describes atmosphere as an emergent affective quality of collective situations, shaped by situational context and interaction~\citep{anderson2009a}.
Drawing on this perspective, we treat dialogue-level affective atmosphere as a latent affective tendency within conversational context.
Compared with TraceERC's use of predicted emotional context as contextual exemplars~\citep{xue2026}, our framework estimates a compact dialogue-level affective prior from a relation-aware graph and evaluates its use in both lightweight sequential decoding and prompt-based ERC.
\newtcolorbox{promptbox}[1]{
    enhanced,
    breakable,
    colback=white,
    colframe=black,
    colbacktitle=black,
    coltitle=white,
    title={#1},
    fonttitle=\bfseries\normalsize,
    boxrule=1.2pt,
    arc=2mm,
    left=2mm,
    right=2mm,
    top=2mm,
    bottom=2mm,
    titlerule=0pt
}

\section{Prompt Design}\label{apx:prompt}

Here we present a schematic prompt template used for the atmosphere plug-in experiments.
For each target utterance, we prepend the verbalized atmosphere descriptor $\widehat{z}$ to the original ERC prompt as a dialogue-level affective prior.
The descriptor $\widehat{z}$ is predicted from the extracted atmosphere vector $\bm{a}$ and does not rely on gold utterance labels during inference.
Apart from this additional atmosphere instruction, the dialogue context, target utterance, candidate emotion set, and decoding procedure follow the original LLM-based ERC method.

\begin{promptbox}{Prompt Template}
You are given a conversation and a target utterance.
The overall affective atmosphere of this conversation is \texttt{[ATMOSPHERE]}.\\
\textbf{Dialogue Context:}\\
\colorbox{mygreen!15}{\texttt{[Dialogue context]}}\\
\textbf{Target Utterance:}\\
\colorbox{myred!15}{\texttt{[Target utterance]}}\\
Please predict the emotion label of the target utterance from the predefined emotion set.
\end{promptbox}
\noindent Here, \texttt{[ATMOSPHERE]} is filled with the textual emotion name corresponding to $\widehat{z}$.
\section{Detailed Experimental Setups}

\subsection{Datasets}\label{app:dataset}

We evaluate our model on three widely used benchmark datasets for ERC. 

\noindent\textbf{IEMOCAP}~\citep{busso2008} is a multimodal ERC dataset consisting of dyadic conversations performed by professional actors in both scripted and improvised scenarios.
Each utterance is annotated with one of six emotion categories: neutral, happiness, sadness, anger, frustrated, and excited.

\noindent\textbf{MELD}~\citep{poria2019} is a multimodal conversational emotion dataset collected from the TV series \emph{Friends}.
It contains multi-party conversations annotated with seven emotion labels: neutral, happiness, surprise, sadness, anger, disgust, and fear.

\noindent\textbf{EmoryNLP}~\citep{zahiri2017} is another ERC dataset derived from TV show scripts of \emph{Friends}, differing from MELD in both scene selection and emotion annotation scheme.
It includes seven emotion categories: neutral, sad, mad, scared, powerful, peaceful, and joyful.

\begin{figure}[!t]
    \centering
    \includegraphics[width=\columnwidth]{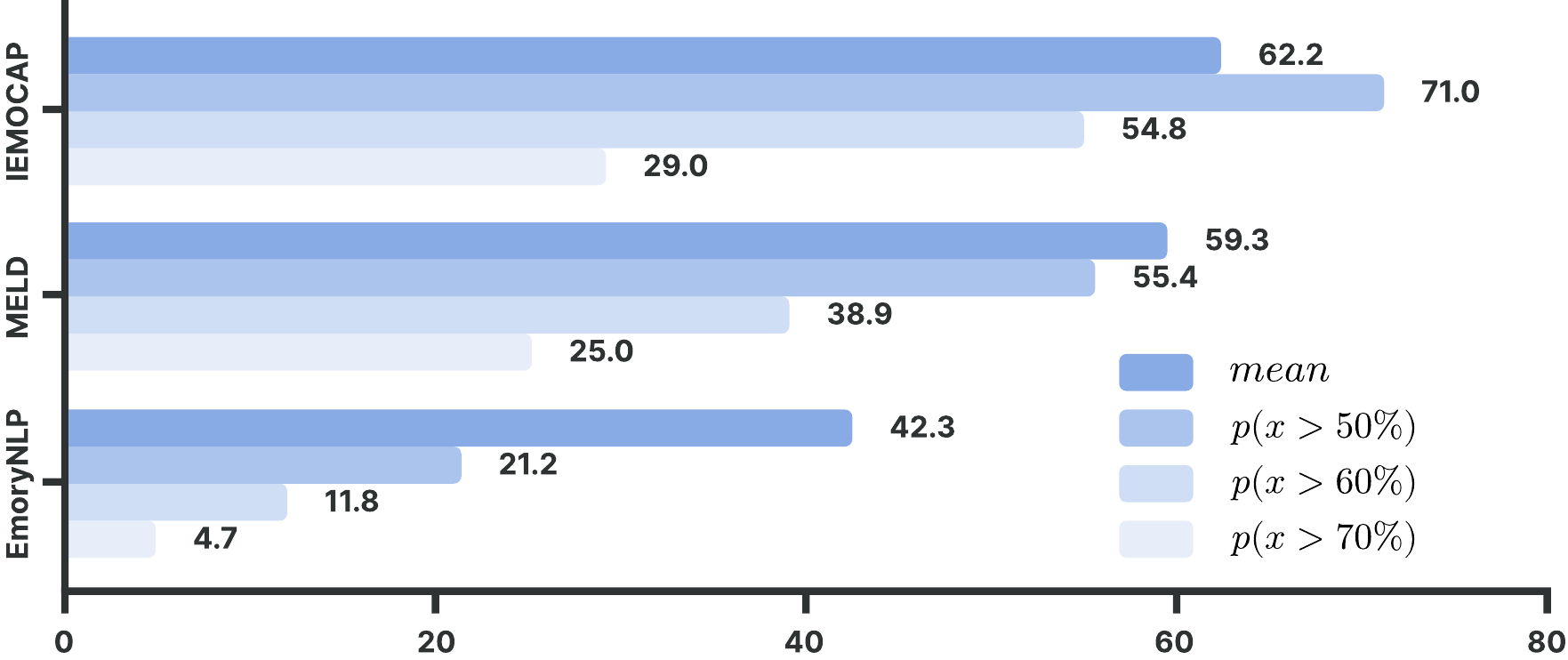}
    \caption{Dominant emotion statistics across three datasets.}
    \label{fig:atmos_statis}
\end{figure}

\noindent We further investigate statistical regularities in utterance-level emotions across three ERC benchmarks.
Figure~\ref{fig:atmos_statis} shows that a single emotion accounts for more than half of the utterances in 71.0\%, 55.4\%, and 21.2\% of dialogues in IEMOCAP, MELD, and EmoryNLP, respectively, revealing substantial differences in dominant-emotion concentration across datasets.
\subsection{Baselines}\label{app:baselines}

We provide detailed descriptions of the baselines used in two evaluation settings: non-LLM methods for lightweight ERC benchmarking and LLM-based methods for evaluating the effectiveness of our prompt-level plug-in.

\paragraph{Non-LLM-based Methods.}
\textbf{SGED}~\citep{bao2022} enhances sequence-based ERC by modeling speaker-aware emotional dynamics with gated recurrent architectures.
\textbf{SACL-LSTM}~\citep{hu2023} incorporates supervised adversarial contrastive learning into an LSTM-based framework to improve emotion representation learning.
\textbf{DAG-ERC}~\citep{shen2021} employs directed acyclic graphs to capture causal and temporal dependencies among utterances.
\textbf{SUNET}~\citep{song2023} constructs a heterogeneous speaker–utterance graph with tailored node updates to capture conversational context, temporal dependencies, and global speaker characteristics for ERC.
\textbf{FrameERC}~\citep{li2025d} employs graph framelet transforms to capture low- and high-frequency emotional cues and introduces dual-reminder fusion to enhance non-textual modality contributions in multimodal ERC.
\textbf{CauAIN}~\citep{zhao2022} introduces causal-aware knowledge modeling to support emotion recognition in dialogue contexts.
\textbf{SKIER}~\citep{li2023} leverages commonsense knowledge to improve contextual emotion understanding.
\textbf{LECM}~\citep{lu2025b} integrates causal emotion entailment as an auxiliary task and employs knowledge-enhanced representations with a cause-oriented tendency perception network to capture emotion-triggering context for real-time ERC.
\textbf{CFN-ESA}~\citep{li2024c} introduces emotion-shift awareness into a cross-modal fusion framework for dialogue emotion recognition.
\textbf{EMART}~\citep{seong2026} employs text-centered, audio-referred multimodal fusion and emotion-wheel-guided supervised contrastive learning to capture complementary acoustic cues and model affective proximity in ERC.
\textbf{EACL}~\citep{yu2024} leverages pretrained contextual representations with contrastive learning objectives for conversational emotion recognition.
\textbf{SEL+SACL}~\citep{lai2026} augments SACL with self-supervised, situation-aware soft emotion labels derived from contextual information and prior human knowledge.

\paragraph{LLM-based Methods.}
\textbf{InstructERC}~\citep{lei2024} is a generative LLM-based ERC framework that reformulates ERC with instruction tuning, retrieval augmentation, and multi-task supervision.
\textbf{GraphEMO}~\citep{tu2026a} integrates multimodal emotion recognition, emotion-cause reasoning, and emotional support within a unified MLLM, leveraging residual task-specific priors and dual graph attention to model hierarchical task dependencies and intra- and cross-modal speaker interactions.
\textbf{Causal-ERC}~\citep{jing2026} integrates multimodal information, models speaker-aware context, and employs causal prompts based on the Peak-End Rule to enhance long-context modeling.
\textbf{TraceERC}~\citep{xue2026} integrates conversational context, LLM-derived speaker personality traits, and dynamic emotional cues to construct speaker-aware, emotion-labeled demonstrations for enhanced ERC.

\subsection{Reproducibility}\label{app:hyper}

We summarize the hyperparameter settings in Table~\ref{tab:hyperparam}.
For all experiments, utterance representations are extracted using RoBERTa, which is used solely as a frozen feature extractor and is not updated in any stage of our framework.
The graph-based atmosphere extractor uses two GAT layers, and the semantic similarity threshold $\tau_s$ for constructing $r_3$ edges is selected on the validation set.
During lightweight ERC training, the atmosphere extractor contains
90M trainable parameters, and the full \method model contains
110M trainable parameters, excluding the frozen RoBERTa encoder.

\begin{table}[h]
\centering
\footnotesize
\setlength{\tabcolsep}{2pt}
\begin{tabular*}{\columnwidth}{@{\extracolsep{\fill}}lccc}
    \toprule
    \textbf{Parameter} 
    & \textbf{IEMOCAP} 
    & \textbf{MELD} 
    & \textbf{EmoryNLP} \\
    \midrule
    Window size $W$     & 4 & 1 & 1 \\
    Threshold $\tau_s$ & 0.95 & 0.95 & 0.90 \\
    GNN layers $L$   & 2 & 2 & 2 \\
    \midrule
    Dropout rate     & 0.1 & 0.1 & 0.2 \\
    Optimizer        & AdamW & AdamW & AdamW \\
    Learning rate    & $5e-5$ & $5e-5$ & $5e-5$ \\
    Weight decay     & $1e-2$ & $1e-2$ & $1e-2$ \\
    Batch size       & 16 & 64 & 32 \\
    Epochs           & 150 & 80 & 50 \\
    \bottomrule
\end{tabular*}
\caption{Hyperparameter settings.}
\label{tab:hyperparam}
\end{table}

For lightweight ERC evaluation, we report baseline results from the original papers when available.
For \method, we report the average performance over multiple random seeds.
For LLM-based plug-in evaluation, we conduct paired comparisons by reproducing each baseline and its atmosphere-augmented variant under the same settings.
For atmosphere verbalization, we hold the trained atmosphere extractor fixed and optimize only the verbalization head using dominant-emotion proxy targets from training dialogues.
The atmosphere descriptor is the only additional input in the augmented variant.
Because exact reproduction of independently reported LLM results is not always possible without complete implementation details, our LLM-based results focus on within-setting comparisons rather than direct comparison with numbers reported in the original papers.
Due to the high cost of LLM inference, each LLM-based method is evaluated with one deterministic decoding pass, where the temperature is set to 0 to reduce sampling variance.
All experiments are conducted on the same machine with two Intel\textregistered Xeon\textregistered Platinum 8457C CPUs, and 8 NVIDIA GeForce RTX 5090 GPUs with 32GB memory each.

\section{Extended Analysis}\label{apx:more_result}

\subsection{Evaluation of Atmosphere Estimation}\label{app:proxy_atmosphere}

We examine the quality of atmosphere estimation through two proxy-based diagnostics.
Since the datasets lack direct atmosphere annotations, we use each dialogue's dominant emotion, which is defined as its most frequent utterance-level emotion label, as an indirect proxy for the overall affective tendency, to compare the estimated representation $\bm{a}$.

\paragraph{Agreement with the Dominant-Emotion Proxy.}
Following the verbalization procedure, we map the estimated atmosphere vector $\bm{a}$ to a predicted dialogue-level emotion descriptor and compare it with the gold-derived proxy.
In this diagnostic, gold-derived proxy labels are used only for evaluation.
Table~\ref{tab:predicted_atmosphere_split} reports agreement rates of $96.77\%$, $82.14\%$, and $60.00\%$ on IEMOCAP, MELD, and EmoryNLP, respectively.
The lower proxy agreement on EmoryNLP than on IEMOCAP suggests that atmosphere estimation may be less accurate on EmoryNLP.
This difference may partly explain the smaller overall performance gains on EmoryNLP in Tables~\ref{tab:main_result} and~\ref{tab:llm_plugin}, as less accurate atmosphere estimates may provide weaker guidance for emotion recognition.

\begin{table}[h]
\centering
\footnotesize
\setlength{\tabcolsep}{3pt}
\begin{tabular*}{\columnwidth}{c@{\extracolsep{\fill}}cc}
    \toprule
    \textbf{IEMOCAP} & \textbf{MELD} & \textbf{EmoryNLP} \\
    \midrule
    96.77 & 82.14 & 60.00 \\
    \bottomrule
\end{tabular*}
\caption{Agreement (\%) between predicted atmosphere descriptors and gold-derived dominant-emotion proxies.}
\label{tab:predicted_atmosphere_split}
\end{table}

\paragraph{Potential for Improvement.}
To examine the potential downstream benefits of more accurate proxy prediction, we replace the predicted atmosphere descriptor with the gold-derived dominant-emotion proxy while keeping the remaining LLM evaluation settings unchanged.
Gold-derived descriptors are used only in this diagnostic and are not available to the standard inference pipeline.
As shown in Table~\ref{tab:proxy_atmosphere}, this replacement further improves weighted-average F1 for both DeepSeek-v4.1-flash and Causal-ERC on all three datasets, suggesting that improving \method's atmosphere estimation and verbalization could bring further gains in LLM-based ERC.

\begin{table}[h]
\centering
\footnotesize
\setlength{\tabcolsep}{2pt}
\begin{tabular*}{\columnwidth}{l@{\extracolsep{\fill}}lll}
    \toprule
    \textbf{Method} & \textbf{IEMOCAP} & \textbf{MELD} & \textbf{EmoryNLP} \\
    \midrule
    DeepSeek-v4.1-flash & 56.99 & 67.18 & 38.05 \\
    ~+ \method-P  
        & 60.01 \gain{604}{+3.02}
        & 67.95 \gain{154}{+0.77}
        & 38.90 \gain{170}{+0.85} \\
    ~+ Gold proxy 
        & 60.66 \gain{734}{+3.67}
        & 69.05 \gain{374}{+1.87}
        & 40.08 \gain{406}{+2.03} \\
    \midrule
    Causal-ERC & 69.13 & 68.10 & 40.64 \\
    ~+ \method-P
        & 71.83 \gain{540}{+2.70}
        & 69.18 \gain{216}{+1.08}
        & 40.94 \gain{60}{+0.30} \\
    ~+ Gold proxy 
        & 72.05 \gain{584}{+2.92}
        & 70.39 \gain{458}{+2.29}
        & 41.54 \gain{180}{+0.90} \\
    \bottomrule
\end{tabular*}
\caption{Comparison results between predicted atmosphere descriptors and gold-derived dominant-emotion proxies.}
\label{tab:proxy_atmosphere}
\end{table}

In summary, the agreement results and gold-proxy gains suggest room for improvement in atmosphere estimation and verbalization.
More accurate prediction of the dialogue-level affective proxy may provide additional benefits for LLM-based ERC.

\subsection{Affective Atmosphere Diagnostics} \label{apx:global_vs_atmos}

To make the comparison between atmosphere and context operational, we define global context as a non-structured pooling of frozen RoBERTa utterance representations.
Specifically, for a dialogue with utterance representations $\{\bm{x}^{(u)}_i\}_{i=1}^{N}$, we compute its context vector as
\begin{equation}
    \bm{c}=\frac{1}{N}\sum_{i=1}^{N}\bm{x}^{(u)}_i.
\end{equation}
Although other context encoders are possible, this choice provides a strong and controlled textual context proxy.
Moreover, the learned atmosphere prior $\bm{a}$ is derived from the same RoBERTa features through relation-aware graph filtering.
Thus, the following analyses compare context pooling $\bm{c}$ with a dominant-emotion-based atmosphere proxy $\bm{p}$ under the same base representation.

\paragraph{Distance-space Analysis.}
We first revisit the distance visualization in Figure~\ref{fig:gct_gea}.
Since true atmosphere is latent, we construct an affective proxy vector $\bm{p}$ by averaging the representations of utterances expressing the dialogue's dominant emotion $z^\star$, and compare it with the context vector $\bm{c}$ defined above.

Specifically, for each dialogue pair $(\mathcal{C}_i, \mathcal{C}_j)$, we compute the global context distance and atmosphere distance by cosine distance:
\begin{equation}
\begin{aligned}
    d_c(i,j) &= 1-\cos(\bm{c}_i,\bm{c}_j), \\
    d_a(i,j) &= 1-\cos(\bm{p}_i,\bm{p}_j).
\end{aligned}
\end{equation}
The overall upward trend in Figure~\ref{fig:gct_gea} shows that atmosphere-proxy distance increases with context distance, indicating that atmosphere remains grounded in context.
However, the large dispersion around the trend line shows that context distance does not fully determine atmosphere-proxy distance.
More importantly, dialogue pairs sharing the same dominant emotion are concentrated in low atmosphere-distance regions and tend to lie below the trend line.
This indicates that affectively similar dialogues can remain close in atmosphere space even when their context differs, supporting the view that atmosphere captures an affect-oriented abstraction rather than simply duplicating global context.

\subsection{Edge Type Proportions}\label{app:edge_ratio}

To clarify the edge type distribution of the constructed graphs, we report the average proportion of each edge type under a representative setting ($W=1$, $\tau_s=0.95$), as shown in Table~\ref{tab:edge_ratio}.
Since the proportions depend on graph construction hyperparameters, these statistics should be interpreted as representative examples rather than fixed dataset properties.
Importantly, the contribution of each relation is not strictly proportional to its frequency.
For example, although $r_3$ accounts for 79.81\% of edges in IEMOCAP, removing it only slightly degrades performance; conversely, in EmoryNLP, where $r_3$ accounts for only 5.30\% of edges, removing it still harms performance.

\begin{table}[h]
\centering
\footnotesize
\setlength{\tabcolsep}{2pt}
\begin{tabular*}{\columnwidth}{l@{\extracolsep{\fill}}ccc}
    \toprule
    \textbf{Relation} & \textbf{IEMOCAP} & \textbf{MELD} & \textbf{EmoryNLP} \\
    \midrule
    $r_1$ & ~~6.17  & 20.18 & 32.41 \\
    $r_2$ & ~~6.92  & 21.31 & 27.27 \\
    $r_3$ & 79.81 & 28.68 & ~~5.30  \\
    $r_4$ & ~~7.12  & 29.83 & 35.02 \\
    \bottomrule
\end{tabular*}
\caption{Average proportions ($\%$) of each edge type under $W=1$ and $\tau_s=0.95$.}
\label{tab:edge_ratio}
\end{table}
\section{Use of AI Assistants}
AI assistants were used only for language polishing, formatting assistance, and consistency checking. 
All research ideas, experimental designs, implementations, analyses, and conclusions were developed and verified by the authors.

\end{document}